\pdfoutput=1

\documentclass[11pt]{article}

\usepackage[preprint]{acl}

\usepackage{times}
\usepackage{latexsym}
\usepackage{graphicx}
\usepackage{booktabs}
\usepackage{ragged2e}
\usepackage{tabularx}

\usepackage[T1]{fontenc}

\usepackage[utf8]{inputenc}

\usepackage{microtype}

\usepackage{inconsolata}

\usepackage{graphicx}

\title{Document Classification using File Names}

\author{Zhijian Li ~~~ Stefan Larson ~~~ Kevin Leach \\
         Vanderbilt University\\
         \texttt{\{firstname.lastname\}@vanderbilt.edu}}

\begin{document}
\maketitle
\begin{abstract}
Rapid document classification is critical in several time-sensitive applications like digital forensics and large-scale media classification.  
Traditional approaches that rely on heavy-duty deep learning models fall short due to high inference times over vast input datasets and computational resources associated with analyzing whole documents. 
In this paper, we present a method using lightweight supervised learning models, combined with a TF-IDF feature extraction-based tokenization method, to accurately and efficiently classify documents based solely on file names, that substantially reduces inference time.
Our results indicate that file name classifiers can process more than 90\% of in-scope documents with 99.63\% and 96.57\% accuracy when tested on two datasets, while being 442x faster than more complex models such as DiT. 
Our method offers a crucial solution to efficiently process vast document datasets in critical scenarios, enabling fast and more reliable document classification.
\end{abstract}

\section{Introduction}

Classifying massive amounts of documents at scale is often an important yet challenging task, especially at industrial scale, where applications of document classification include sensitive data identification, legal document review, spam and malware flagging, etc.
Large-scale classification systems often process millions of documents, requiring robust models to handle diverse taxonomies efficiently~\cite{kandimalla2021large}.
In sensitive data identification, document sensitivity risk scores can be assigned according to its category --- for instance, academic transcripts and medical intake forms may be marked sensitive, while press releases may be marked low-risk.

Typical approaches to document classification use text-, image-, layout-, and multimodal classifiers (e.g., \citet{li2022dit, layoutlmv3-huang-2022, longformer-2022-pham}), but these models can be prohibitively slow at the scale required in many industry settings~\cite{park-etal-2022-efficient}, 
as these models must execute sophisticated operations on the contents of a very large quantity of documents.
Thus, there is a need for techniques that speed up the document classification pipeline to enable scale.

\begin{figure}
    \centering\scalebox{0.385}{
    \includegraphics{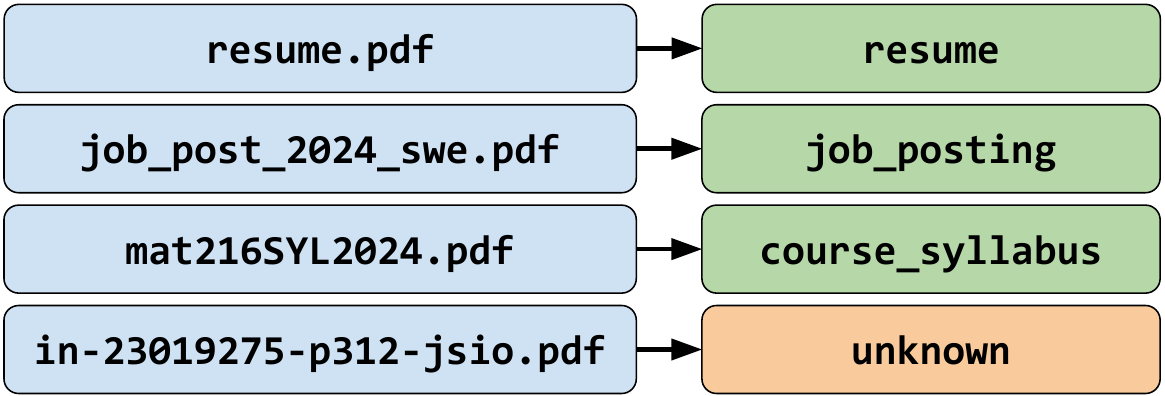}}
    \caption{Overview of the file name classification task. An ideal classifier can categorize indicative file names correctly, while also being able to defer predictions on file names that lack enough category information.}
    \label{fig:teaser}
\end{figure}

In this paper, we introduce and investigate the task of classifying documents by categories (e.g., \texttt{resume}, \texttt{press\_release}, etc.) using only their file names as input.
Figure~\ref{fig:teaser} displays several examples where documents can be categorized based on their file names.
An added challenge is to correctly identify cases that are \emph{ambiguous}, which are documents whose file names do not provide a clear indication of their content.
However, since many file names are \emph{indicative} of their contents (e.g., \texttt{john\_resume\_2024.pdf} $\rightarrow$ \texttt{resume}), we hypothesize that much simpler classifiers can be trained and used to categorize these documents.  
Furthermore, we hypothesize that a sufficiently large portion of documents can be classified using file name alone to yield a substantial savings in compute resources and time.

Our experimental results show that the Random Forest file name classifier with a trie tokenizer achieves 99.63\% and 96.57\% accuracy on two respective datasets while deferring less than 10\% of the in-scope documents.
This Random Forest file name classifier requires only $1.23 \times 10^{-4}$ seconds per prediction, which is 442x faster than DiT~\cite{li2022dit}, an exemplar image-based deep learning model that requires access to each document's content which contains both images and extracted text. 

\begin{table*}[]
    \centering
    \scalebox{0.97}{
    \begin{tabular}{lcl}
    \toprule
         \textbf{File Name} & \textbf{Ambiguity Type} & \textbf{Category} \\
    \midrule
         \texttt{Cerno-Woodshop-Assistant-Opportunity.pdf} & indicative & \texttt{job\_posting} \\
         \texttt{TcallsenResume2020v0710-aem-react.pdf} & indicative & \texttt{resume} \\
         \texttt{The-Willow-Restaurant-Menu-0422.pdf} & indicative & \texttt{restaurant\_menu} \\
         \texttt{StJudeNewsRelease2011.pdf} & indicative& \texttt{press\_release} \\
         \texttt{MX\_7090N\_8090N\_Spec\_Sheet.pdf} & indicative& \texttt{specification} \\
         \texttt{dear-colleague-letter-fda-samhsa.pdf} &indicative & \texttt{letter} \\
         \texttt{BACB\_September2023\_Newsletter-231201-a.pdf} & indicative& \texttt{newsletter} \\
         \midrule
         \texttt{12042023-NMPH.pdf} & ambiguous& \texttt{job\_posting} \\
         \texttt{plnbudm3-30-4-6s12.pdf} & ambiguous & \texttt{meeting\_minutes} \\
         \texttt{5bba5a94c8f1454f98d7b01be13289b9.pdf} & ambiguous& \texttt{policy} \\
         \texttt{privacy-notice.pdf} & out-of-scope & \texttt{privacy\_notice} \\
         \texttt{PremilinaryBalance\_2011.pdf} & out-of-scope& \texttt{financial\_sheet} \\
         
    \bottomrule
    \end{tabular}}
    \caption{Examples of indicative file names, ambiguous, and out-of-scope file names.}
    \label{tab:examples_file_names}
\end{table*}

\section{Related Work}
In this section, we discuss prior work on classifying files using file names, and document classification.

\subsection{Classification with File Names}
Document classification via file names is an understudied topic.
While prior research on categorizing content via file names has shown promise, it has largely focused on binary classification.
For instance, \citet{alnabki-2023-short-text-classification-cse} proposed an approach to classify  Child Sexual Exploitation Material (CSEM) through analyzing the file names and file paths of files when law enforcement officers seize a computer from a CSEM suspect due to the time-consuming nature of heavy weight models.
Similarly, \citet{2021-spam-email} accurately detected spam emails through analyzing email names using a Naive Bayes classifier, and \citet{malware-filename-nguyen-2019} demonstrated that a character-level CNN can accurately classify file names that contain enough information indicative of malware. 
Additionally, \citet{url-phishing-detection-2022} accurately detected phishing URLs using a logistic regression model combined with Token Frequency-Inverse Document Frequency (TF-IDF) feature extraction.

These studies demonstrate that binary classification through file names or document titles alone can be effective without deeper analysis of the contents of the files themselves.
However, multi-class file name classification tasks, such as document classification by categories, remains under-studied. 
Given the demand for efficient large-scale document processing in industry settings~\cite{8768370}, further research in this area is needed.
Moreover, the issue of handling data that fall outside the scope of the classifiers or those that lack clear indicators for accurate classification illustrated in~\cite{larson-etal-2019-evaluation} has not yet been adequately addressed in the existing literature for document classification.

\subsection{Document Classification}
Prior work on document classification includes the development of large image-, text-, and page layout-based models for categorizing documents, mostly from datasets derived from the IIT-CDIP tobacco corpus (e.g., RVL-CDIP \cite{harley2015icdar} and Tobacco3482 \cite{tobacco3482-2013-kumar}). 
These large models include the LayoutLM family (LayoutLM \cite{xu-layoutlm-2020}, LayoutLMv2 \cite{xu-etal-2021-layoutlmv2}, LayoutLMv3 \cite{layoutlmv3-huang-2022}), DiT \cite{li2022dit}, DocFormer \cite{Appalaraju_2021_ICCV-docformer}, Donut \cite{donut-kim-2022-eccv}, and LayoutLLM \cite{fujitake-2024-layoutllm}, etc.
However, these models are costly to train and costly to run.
Image-based models like CNNs and Transformers must operate on images of document pages, and text- and layout-based models must run on variants of the document from which text has been extracted via Optical Character Recognition (OCR), which adds a layer of complexity for each document that must be classified.



\section{Problem Overview}

In this study, we examine whether we can efficiently and accurately classify documents only based on their file names. Table~\ref{tab:examples_file_names} shows examples of file names, their ambiguity types, and their categories.
To enhance the clarity of our analysis, we distinguish the file names into the three following ambiguity types:
\begin{itemize}
    \item \textbf{Indicative:} File names that contain the category name, a synonym of that category, or an abbreviation of the category name. Examples of indicative file names include \texttt{john\_smith\_resume\_2024.pdf} (a \texttt{resume} document), \texttt{john\_smith\_cv\_2024.pdf} (a \texttt{resume} document, where ``cv'' is used as a synonym for ``resume''), and \texttt{RTKRequestForm.pdf} (a \texttt{form} document).

    \item \textbf{Ambiguous:} File names that do \emph{not} contain the category name itself or a synonym of the category name, but belongs to a category in the dataset. Examples of ambiguous file names include \texttt{20-005-10.pdf} and \texttt{document.pdf}, which have names that are not indicative, but the content may happen to indicate the document belongs to a particular category in the dataset. 
    
    \item \textbf{Out-of-Scope:} File names that do \emph{not} belong to any category in the dataset. We note that out-of-scope file names are---to borrow terminology from \citet{larson-etal-2019-evaluation}---``out-of-scope'' with respect to the in-scope or in-domain file categories defined in the dataset.
\end{itemize}


A key aspect of our approach is to determine whether the file name is indicative based on the confidence score associated with the predicted class for that file name. 
In cases where the confidence is too low, we \emph{defer} the file name.
In practice, we would instead use a more complex model at increased computational cost if the file is deferred by the file name classifier.
In cases where the confidence score is high enough, we desire a highly accurate class prediction to forego the additional expense of running a deeper, more complex model involving the file's contents. 
With a higher confidence threshold, the file name classifiers are more accurate with these predictions, but predict a smaller portion of all the file names, ultimately adding computational burden to the heavier-duty model in the next stage. 
We discuss the trade-off and optimal confidence threshold for each file name classifier in Section~\ref{sec: eval}.
We also measure the AUROC of the confidence scores between indicative and ambiguous file names to analyze whether the file name classifiers produce confidence scores that differentiate between indicative and ambiguous or out-of-scope file names.

In this study, we only compare the inference speed of the heavier-duty models as we are interested how much faster the file name classifiers are compared to models such as DiT, rather than trying to outperform the more complex models in accuracy.
This study aims to address whether file name classifiers can accurately predict indicative file names, while \emph{deferring} the ambiguous and out-of-scope file names.
If a sufficiently large portion of documents contain indicative file names, then the savings of using a lightweight classifier before running a heavy duty model will substantially outweigh the cost of using a heavy duty model on the entire dataset.

\section{Method}
In this section, we discuss our approach to leveraging indicative and ambiguous file names to develop a fast, accurate document classification scheme that reduces the overall computational resources required to classify large volumes of documents. 
Our approach consists of a tokenization scheme for splitting input file names as well as training and evaluating a light-weight machine learning pipeline. 



\subsection{Tokenization}\label{sec: tokenization}

\begin{figure}
    \centering\scalebox{0.5}{
    \includegraphics{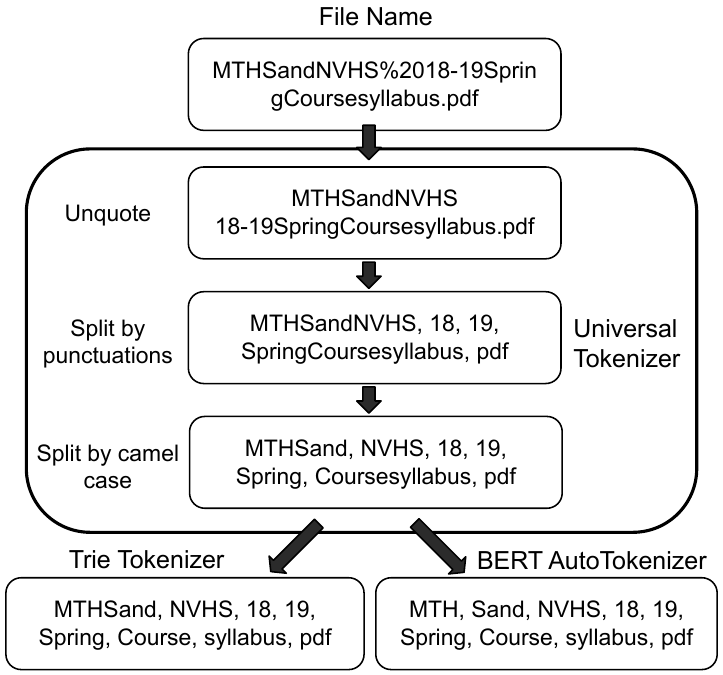}}
    \caption{An example of the universal tokenizer tokenizes a fileame with the trie tokenizer and BERT Autotokenizer.}
    \label{fig:appendix_overview}
\end{figure}

We apply a ``universal tokenizer" to clean all input file names. 
The ``universal tokenizer" first preprocess each file name by replacing escape sequences (e.g., ``\%xx'') with the relevant single-character equivalent. 
We then split the file names by all punctuation characters, white space, camel case boundaries (i.e., when a lower-case character is immediately followed by an upper-case character), and transitions between letters ([a-zA-Z]) and numbers ([0-9]).

The ``universal tokenizer" is not equipped to handle cases where file names contain words joined together without any punctuation or camel case such as \texttt{coursecatalog.pdf}.
From randomly sampling 1,000 file names and annotating the performance of the ``universal tokenizer", we discover that 3.4\% file names needs further tokenization.
Figure~\ref{fig:appendix_overview} shows an example where the ``universal tokenizer" fails to separate \texttt{Coursesyllabus} into ``Course'' and ``syllabus,'' instead requiring additional tokenization.

To improve the tokenization for such file names, we first calculate the Token Frequency-Inverse Document Frequency of every token in relation to the category. 
After lemmatizing the output tokens with NLTK version 3.6.7's WordNetLemmatizer~\cite{bird-loper-2004-nltk}, tokens with TF-IDF higher than a threshold $k$ are identified as ``keywords.''
We choose $k$ to be 0.2 in this study. 
The trade-offs of selecting different value of $k$ is discussed in Section~\ref{sec: k value}. 
The keywords are then added to a trie data structure to create a trie tokenizer that can effectively locate and split important words that are joined together if the words in the file names are not separated with special characters, whitespace, or camel case. 

Another approach involves adding the ``keywords'' to a BERT tokenizer~\cite{wolf-etal-2020-transformers} using the ``add\_tokens'' function in the Hugging Face library.
For example, the trie or BERT tokenizer will split file names such as ``coursecatalog'' in ``course'' and ``catalog'' to enable effective bag-of-words encoding.

This leaves us with a set of tokens for each file name, allowing us to find indicative cases where the file name makes a clear and straightforward hint as to the underlying document's class.

\subsection{Classification}
We use the following text classifiers in our experiments:

\textbf{Support Vector Machine (SVM)}, a supervised learning model that classifies data by finding a separating hyperplane that maximizes the margin. 
We use the linear kernel and the regularization parameter $c=0.1$ as this combination yields the highest accuracy during our hyperparameter search where we compare the accuracy on the SVMs trained and tested on indicative file names with different kernels and value of $c$. 
The SVM model is implemented with Scikit-Learn version 1.4.2~\cite{pedregosa2011scikit} which uses a one-vs.-all approach. 

\textbf{Naive Bayes}, a supervised learning model that learns the distribution of inputs by assuming that all the features are independent of each other. 
We implement this model with Scikit-Learn \cite{pedregosa2011scikit}.

\textbf{Random Forest}, a supervised learning model that aggregates the prediction of many decision trees to prevent over-fitting.
We implement this model with 100 decision trees through Scikit-Learn \cite{pedregosa2011scikit}.

\textbf{eXtreme Gradient Boosting (XGBoost)}, a supervised learning model that is implemented with gradient boosted decision trees. 
It is implemented with the XGBoost library~\cite{Chen:2016:XST:2939672.2939785}.

\textbf{BERT}, a neural network based on a Transformer model, capture the context of each word \cite{devlin-etal-2019-bert}.  
In this study, we fine-tune the BERT-base-uncased model from Hugging Face\footnote{\url{https://huggingface.co/google-BERT/BERT-base-uncased}} on the Web Search dataset to classify file names.

When preprocessing the file names, we first use the tokenization technique discussed in Section~\ref{sec: tokenization}. 
Subsequently, for the SVM, Naive Bayes, Random Forest, and XGBoost classifiers, we encode the file name tokens with a bag-of-words encoding.
For BERT, we join the tokens with space and then encode the string with the BERT tokenizer.

\subsection{Metrics}\label{sec:metrics}

For a selected confidence threshold $t$, we analyze the \textbf{prediction accuracy} by calculating $\frac{\#\mathrm{correct}}{\#\mathrm{correct} + \#\mathrm{incorrect}}$,
the number of correct predictions divided by the number of file names whose prediction confidence exceeds $t$. 
We examine the \textbf{prediction rate} by calculating
$\frac{\#\mathrm{correct} + \#\mathrm{incorrect}}{\mathrm{total}\#\mathrm{documents}}$, the sum of the correct and incorrect predictions over the total number of file names.
Prediction Accuracy over 0.9 prediction rate is calculated to evaluate the models' abilities to leverage confidence scores to avoid erroneous predictions while still processing the majority of the data.

Additionally, to analyze the models' ability to distinguish ambiguous file names from the indicative file names, we calculate the confidence score for each prediction. 
For BERT, we use the highest softmax confidence score for each prediction~\cite{hendrycks2016baseline}. 
For SVM, Naive Bayes, and Random Forest, we use the highest value from the \texttt{predict\_proba} function through Scikit-Learn.
Similar to \citet{larson-2022-rvlcdip-oos}'s work which used the Area Under Receiver Operating Characteristic curve (AUROC) to evaluate the models' ability to distinguish out-of-scope samples from in-scope samples, we use AUROC to measure how well each model can distinguish ambiguous from indicative file names via confidence scores .


\section{Datasets} \label{dataset_section}

We consider two datasets used as part of our evaluation.

\subsection{Web Search Dataset} \label{dataset: ws}
We collect a dataset with 1,500 PDF files with indicative file names from 15 categories through web searches, each category containing 100 files.
A list of these 15 categories is shown in Table~\ref{tab:appendix-web-search-dataset-categories} in the Appendix.

We consider search prompts of the format:
\begin{Center}
\texttt{<category\_name> filetype:pdf}
\end{Center}
(where \texttt{filetype:pdf} specifies that we desire PDF documents from the web search engine).
For example, we gather the PDF files for restaurant menus by searching 
\begin{Center}
\texttt{restaurant menu filetype:pdf}.
\end{Center}
To obtain file names, we select the substring after the last ``/'' from each document's URL.
Then, we manually open the files to verify that they belong to the correct category, and if their file names are indicative.

We also note that synonyms or related words and phrases are also used when searching for category names.
For instance, when searching for restaurant menus, we also searched for ``diner menu,'' ``bar menu,'' etc.

The Web Search dataset contains 15 categories, which is similar in number to the RVL-CDIP dataset \cite{harley2015icdar} which contains 16 categories.
Table~\ref{tab:examples_file_names} displays more examples of file names from this dataset.

We observe that numerous file names in the ``resume'' category contains people's names. 
We replace the original last names by randomly generating last names using the Faker Python package version 36.1.1~\cite{faker} for each of those file names.

\subsection{Common Crawl Dataset} \label{dataset: cc}
In our evaluation, we also consider another open source dataset to demonstrate generalizability of our approach. 
We use a subset of the CC-MAIN-2021-31 dataset, which consists of PDFs, URLs, and metadata of eight million unlabeled documents collected by Common Crawl from the web in July and August of 2021.\footnote{\url{https://digitalcorpora.org/corpora/file-corpora/cc-main-2021-31-pdf-untruncated/}}
We randomly select 1,500 
documents from the CC-MAIN-2021-31 dataset, and discard 302 documents that do not contain valid URLs, are not in English or results in label disagreements. 
Of these 1,198 file names, there are 284 indicative file names belonging to one of the 15 document categories defined in the Web Search dataset from Section~\ref{dataset: ws}, 264 ambiguous file names, and 650 out-of-scope file names.

The correct category labels and ambiguity types of the Common Crawl dataset are obtained by manual annotation by two annotators. 
When the annotators cannot reach an agreement on the appropriate category, the document is discarded.

\section{Results} \label{sec: eval}

We discuss the experimental results and settings in this section.

SVM, Naives Bayes, Random Forest, XGBoost, and BERT classifiers are trained on 80\% of the indicative file names from the Web Search dataset, and evaluated against the rest of the data from the Web Search dataset and the entire Common Crawl dataset.
We perform 5-fold cross validation and calculate the average accuracy, prediction accuracy, and prediction rate (defined in Section~\ref{sec:metrics}).
The subsections below discuss the accuracy and speed of our file name classifiers, as well as their ability to distinguish ambiguous and out-of-scope data.

\subsection{Model Accuracy for Indicative File Names} \label{sec: robustness}

First, we show that our approach can accurately classify indicative file names.
Table~\ref{websearch_acc_trie_table} and table~\ref{websearch_acc_bert_table} shows the accuracy for all models with the trie and BERT tokenizer on the Web Search dataset respectively. 
The BERT file name classifier achieves the highest accuracy of 95.51 with a trie tokenizer, while Random Forest, Naive Bayes, and SVM also achieves above 94\% accuracy.

Additionally, we show the prediction accuracy with at least 0.9 prediction rate iterating through all confidence thresholds between 1 and 0, decrementing by 0.01.
Predictions that result in a confidence score lower than the threshold are deferred by the file name classifier.
Deferring less than 10\% of the data, the Random Forest classifier with the BERT tokenizer produces the highest accuracy of 99.78\% among all combinations of models and tokenizers.
SVM and BERT also achieved more than 99\% accuracy while predicting more than 90\% of the data, while there is a gap in accuracy for the Naive Bayes model.

In addition, we display the performance of the five models trained on the Web Search dataset and tested on the in-scope, indicative file names from the Common Crawl dataset in table~\ref{cc_acc_trie_table} and table~\ref{cc_acc_bert_table}.
The Random Forest classifier with the trie tokenizer achieves 96.57\% prediction accuracy with above 0.9 prediction rate.
We observe that the Random Forest classifier obtains the highest prediction accuracy among all five models, consistent with the evaluation results from the Web Search dataset.

We observe insubstantial differences in accuracy and prediction accuracy for every file name classifier with different tokenizers.

\begin{table}
    \centering\scalebox{0.7}{
    \begin{tabular}{lrrrr}
        \toprule
        \textbf{Models} & \textbf{Accuracy} & \textbf{Threshold} & \textbf{Prediction Accuracy} \\
        \midrule
        SVM & 94.06 & 0.55 & 99.34   \\
        Naive Bayes & 94.13 & 0.37 & 98.02  \\
        Random Forest & 94.92 & 0.51 & 99.63   \\
        XGBoost & 92.61 & 0.73 & 99.41   \\
        BERT & 95.51 & 0.96 & 99.42   \\
        \bottomrule
    \end{tabular}}
    \caption{The accuracy and prediction accuracy with prediction rate more than 0.9 for all models with the \textbf{trie tokenizer} on the Web Search dataset.}
    \label{websearch_acc_trie_table}
\end{table}

\begin{table}
    \centering\scalebox{0.7}{
    \begin{tabular}{lrrrr}
        \toprule
        \textbf{Models} & \textbf{Accuracy} & \textbf{Threshold} & \textbf{Prediction Accuracy} \\
        \midrule
        SVM & 94.06 & 0.49 & 99.12   \\
        Naive Bayes & 94.39 & 0.35 & 98.17    \\
        Random Forest & 94.72 & 0.50 & 99.78   \\
        XGBoost & 90.76 & 0.6 & 98.97   \\
        BERT & 94.92 & 0.97 & 99.35   \\
        \bottomrule
    \end{tabular}}
    \caption{The accuracy and prediction accuracy with prediction rate more than 0.9 for all models with the \textbf{BERT tokenizer} on the Web Search dataset.}
    \label{websearch_acc_bert_table}
\end{table}

\begin{table}
    \centering\scalebox{0.7}{
    \begin{tabular}{lrrrr}
        \toprule
        \textbf{Models} & \textbf{Accuracy} & \textbf{Threshold} & \textbf{Prediction Accuracy} \\
        \midrule
        SVM & 84.23 & 0.22 & 92.24   \\
        Naive Bayes & 87.96 & 0.25 & 95.00  \\
        Random Forest & 88.24 & 0.31 & 96.57   \\
        XGBoost & 84.93 & 0.30 & 93.69   \\
        BERT & 83.45 & 0.77 & 89.70   \\
        \bottomrule
    \end{tabular}}
    \caption{The accuracy and prediction accuracy with prediction rate more than 0.9 for all models with the \textbf{trie tokenizer} on the indicative documents in the Common Crawl dataset.}
    \label{cc_acc_trie_table}
\end{table}

\begin{table}
    \centering\scalebox{0.7}{
    \begin{tabular}{lrrrr}
        \toprule
        \textbf{Models} & \textbf{Accuracy} & \textbf{Threshold} & \textbf{Prediction Accuracy} \\
        \midrule
        SVM & 84.37 & 0.23 & 92.82   \\
        Naive Bayes & 86.48 & 0.26 & 92.30   \\
        Random Forest & 87.82 & 0.26 & 96.33   \\
        XGBoost & 84.86 & 0.32 & 93.75   \\
        BERT & 82.96 & 0.70 & 89.31   \\
        \bottomrule
    \end{tabular}}
    \caption{The accuracy and prediction accuracy with prediction rate more than 0.9 for all models with the \textbf{BERT tokenizer} on the indicative documents in the Common Crawl dataset.}
    \label{cc_acc_bert_table}
\end{table}

\subsection{Detection Ability for Ambiguous and Out-of-Scope File Names} \label{sec:oos_detect}

We train file name classifiers on the Web Search dataset described in Section~\ref{dataset: ws} and evaluate our models against the Common Crawl dataset described in Section~\ref{dataset: cc}. 
We test our file name classifiers against the ambiguous and out-of-scope file names and evaluate our model's robustness predicting ambiguous file names by calculating the AUROC of the confidence scores between the indicative and ambiguous file names.

\begin{figure}
    \centering\scalebox{0.5}{
    \includegraphics{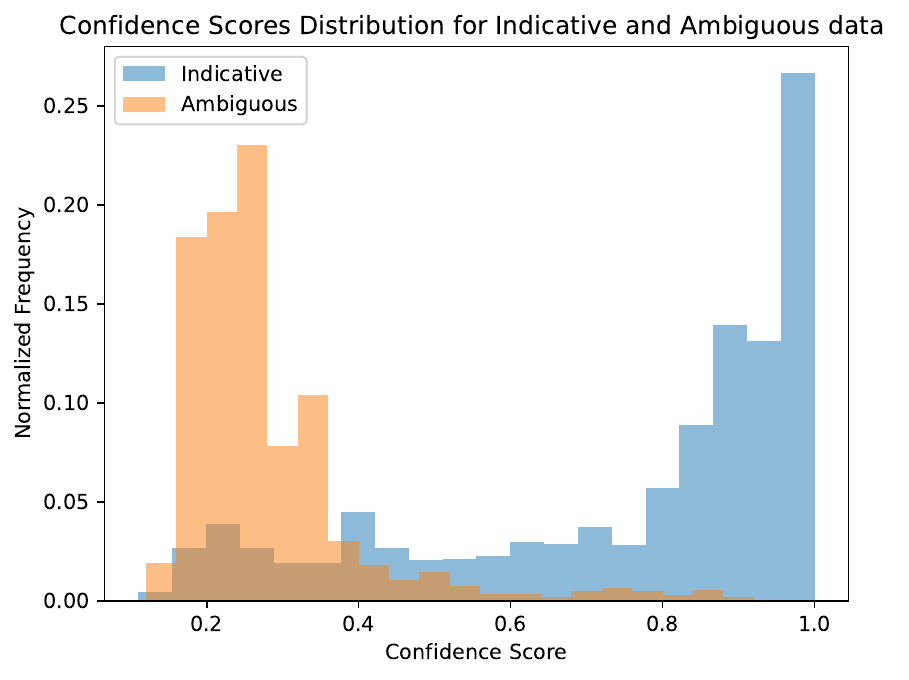}}
    \caption{The confidence score distribution for Random Forest classifier with trie tokenizer on indicative and ambiguous file names from the Common Crawl dataset.}
    \label{fig:rf_confidence_amb}
\end{figure}

Figure~\ref{fig:rf_confidence_amb} shows minor overlap between the confidence scores for the indicative and ambiguous file names for the Random Forest file name classifier with the trie tokenizer.

Table~\ref{table: auroc} displays the AUROC of the confidence scores between indicative and ambiguous file names for all models and each tokenizer. 
The Random Forest file name classifier with the trie tokenizer achieves the highest AUROC of 0.922 which shows that this classifier produces confidence scores that best differentiate indicative and ambiguous file names.
In comparison, Table~\ref{table: auroc_oos} displays that all models output higher AUROC for the indicative and out-of-scope confidence scores, further implying that the Random Forest file name classifier outputs confidence scores that can distinguish most indicative file names from all others.
The differences in AUROC for different tokenizers are also not substantial for all models.

\begin{table}
    \centering\scalebox{0.64}{
    \begin{tabular}{lrrrrr}
        \toprule
        \textbf{Tokenizer} & \textbf{SVM} & \textbf{Naive Bayes} & \textbf{Random Forest} & \textbf{XGBoost} & \textbf{BERT} \\
        \midrule
        trie & 0.894 & 0.890 & 0.922 & 0.905 & 0.868 \\
        BERT & 0.892 & 0.876 & 0.918 & 0.909 & 0.887 \\
        \bottomrule
    \end{tabular}}
    \caption{The AUROC of confidence scores between the indicative and \textbf{ambiguous} file names for all the classifiers with a trie or BERT tokenizer.}
    \label{table: auroc}
\end{table}

\subsection{Inference Time} \label{sec: speed}

\begin{table}
    \centering\scalebox{0.64}{
    \begin{tabular}{lrrrrr}
        \toprule
        \textbf{Tokenizer} & \textbf{SVM} & \textbf{Naive Bayes} & \textbf{Random Forest} & \textbf{XGBoost} & \textbf{BERT} \\
        \midrule
        trie & 0.905 & 0.917 & 0.923 & 0.906 & 0.876 \\
        BERT & 0.909 & 0.915 & 0.925 & 0.912 & 0.886 \\
        \bottomrule
    \end{tabular}}
    \caption{The AUROC of confidence scores between the indicative and \textbf{out-of-scope} file names for all the classifiers with a trie or BERT tokenizer.}
    \label{table: auroc_oos}
\end{table}

Table~\ref{table: inference_time} shows the file name classifiers' average log inference time per prediction for each model on a computer with a single 24-core x84\_64 CPU and no GPUs.

Among all the models, Naive Bayes was the fastest, with a average log inference time of $-13.34$, requiring $1.61 \times 10^{-6}$ seconds per prediction.
The BERT file name classifier's inference time is substantially slower than all four other models, requiring $6.14 \times 10^{-2}$ seconds per prediction.
The Random Forest classifier that achieves the highest prediction accuracy on both the Web Search dataset and the Common Crawl dataset requires $1.30 \times 10^{-4}$ seconds per prediction.

In comparison, when used on the same computer as the file name classifiers, the BERT model that uses the first 512 tokens of a document content as input requires $0.0678$ seconds per document. 
Meanwhile, the DiT model on average requires $0.0574$ seconds per image running on a GPU.
The robust Random Forest file name classifier is 442x faster than the DiT model even if the DiT model only processes the first page of every document and is 552x faster than the BERT model that processes the content of the documents, offering a dramatic speedup.

\begin{table}
    \centering\scalebox{0.74}{
    \begin{tabular}{lrrrr}
        \toprule
         \textbf{SVM} & \textbf{Naive Bayes} & \textbf{Random Forest} & \textbf{XGBoost} & \textbf{BERT} \\
        \midrule
         -10.34 & -13.34 & -8.95 & -10.82 & -2.79 \\
        \bottomrule
    \end{tabular}}
    \caption{Logarithm of average inference time for SVM, Naives Bayes, Random Forest, XGBoost, and BERT file name classifiers, calculated by applying the natural log to the average inference time in seconds.}
    \label{table: inference_time}
\end{table}

\begin{figure}
    \centering\scalebox{0.52}{
\includegraphics{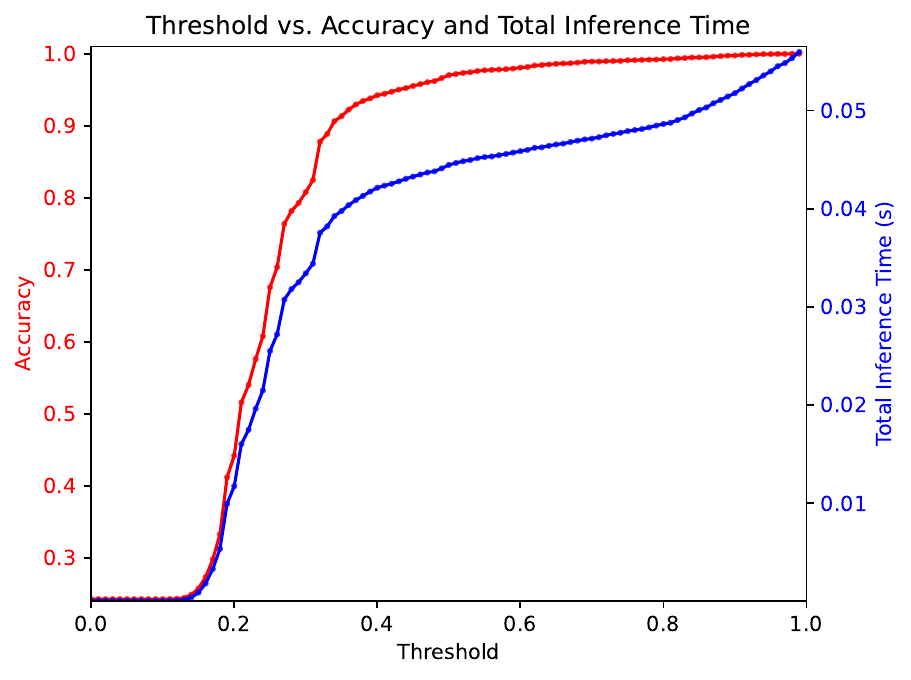}}
    \caption{Overall Accuracy versus Total Inference Time (seconds) 
 per file name at all confidence score threshold between 0 and 1, in increments of 0.001, for a Random Forest classifier with the trie tokenizer predicting on all the file names in the Common Crawl dataset, with a DiT model processing the deferred documents. }
    \label{fig:overall_time_acc_threshold}
\end{figure}

\subsection{Accuracy and Cost Tradeoffs} \label{sec: speed2}
Due to the differences in how we obtain confidence scores for each models, the optimal confidence threshold differs empirically between models. 
Increasing the confidence threshold increases the prediction accuracy while decreasing the prediction rate.
In practice, we must select an appropriate confidence threshold to balance the tradeoff between the prediction accuracy and the fraction of documents accurately predicted as indicative --- the documents that are deferred as ambiguous or out-of-scope would require further examination using a deeper and more complex model. 
Thus, the average time and resource expenditure per document classified would increase with the increase in confidence threshold for predicting a document file name as indicative.

Figure~\ref{fig:overall_time_acc_threshold} shows the accuracy and time inference time per document using the Random Forest file name classifier predicting on all the file names in the Common Crawl dataset, and DiT model processing the deferred documents.
With the assumption that all DiT predictions are correct, accuracy is calculated by prediction accuracy $\times$ prediction rate $+$ ($1-$ prediction rate), and total time is calculated by average inference time of Random Forest + average inference time of DiT * ($1 -$ prediction rate).
The inference time for each model is discussed in Section~\ref{sec: speed}.
At a confidence threshold of 0.31, the improvement in accuracy begins to taper off, while the total time increases more steeply, indicating that an optimal confidence threshold should be chosen close to that value during deployment.
Due to the Common Crawl dataset being randomly sampled from the web without attempting to match the label space of the Web Search dataset, only 23.7\% file names in the Common Crawl dataset are indicative and in-scope. 
Since the majority of the file names are ambiguous or out-of-scope, the file name classifier defers those predictions, which leads to a smaller speedup illustrated in the figure. 
Thus, better alignment between the label spaces of the training and testing data would lead to more drastic speedups in end-to-end performance.

\subsection{Failure Modes} \label{sec:failure_modes}
We analyze the failure modes of the Random Forest file name classifier with the trie tokenizer.
We observe that each class produces close to the same number of errors when tested on the Web Search dataset.
When this file name classifier is tested against the Common Crawl dataset with a confidence threshold of 0.31, we observe that 7.55\% of the ambiguous file names are incorrectly classified, while 12.55\% of the out-of-scope file names are not deferred.

\subsection{Trade-offs for $k$ Value Selection} \label{sec: k value}
\begin{figure}
    \centering\scalebox{0.5}{
    \includegraphics{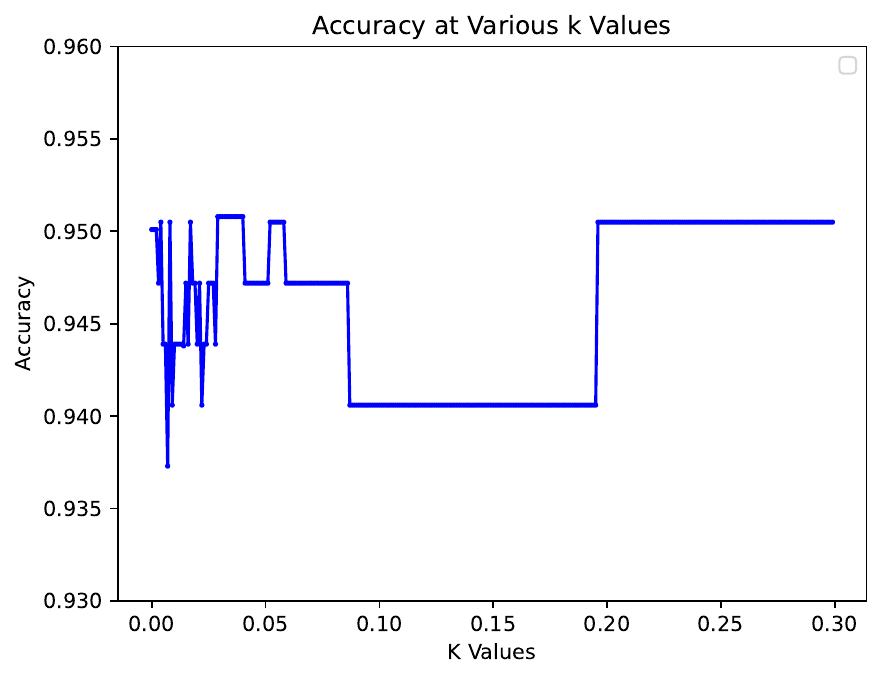}}
    \caption{Accuracy of the Random Forest file name classifier with a trie tokenizer with all k values between 0 and 0.3, incremented by 0.001, tested on the Web Search dataset.}
    \label{fig:k_value}
\end{figure}

In Section~\ref{sec: tokenization}, we mentioned that tokens with TF-IDF larger than $k$ are considered ``keywords'' and added to the tokenizer.
We examine the trade-offs for the selection of $k$ value for the trie tokenizer by comparing the prediction accuracy and prediction rate with various $k$ values.
The previous subsections indicate that the Random Forest file name classifier produces the highest prediction accuracy with a fixed minimum prediction rate.
Figure~\ref{fig:k_value} displays the accuracy and prediction for the Random Forest file name classifiers with $k$ values from 0 to 0.3. 
For $k$ values between 0 and 0.059, the accuracy changes sporadically.
We observe that the accuracy increases substantially when $k$ values reaches 0.196, then remains constant as $k$ values increases.
In this study, we select the $k$ values to be 0.2.

\section{Conclusion}
We present a new approach to multi-class document classification that dramatically improves classification efficiency while achieving high accuracy.
Additionally, we contribute two new document classification datasets that contain file names.
Our evaluation shows that a Random Forest classifier with a trie tokenizer, when tasked to predict at least 90\% of the data, can achieve 99.78\% prediction accuracy on indicative file names from the Web Search dataset and 96.6 \% prediction accuracy on the indicative file names from the Common Crawl dataset.
The Random Forest classifier with a trie tokenizer can also produce confidence scores that can effectively distinguish indicative and ambiguous file names, resulting in an AUROC of 0.922 for confidence scores between indicative and ambiguous.
Meanwhile, the Random Forest classifier is 442x faster than the DiT model, indicating that uses the Random Forest file name classifier can provide a substantial overall speedup in document classification.
We hope that our technique and analysis will enable faster and computationally efficient document classifiers without sacrificing on accuracy.

\section*{Limitations}
In this study, we solely focus on document classification using file names and obtained promising results. We believe metadata --- such as file size, page count, number of images --- can be used along with file names to improve the classifier accuracy without substantial increase in inference time. However, due to time constraints, we are not able to explore this approach. Future work can investigate the potential benefits of combining metadata with file name to improve classification accuracy and handle ambiguous data or out-of-scope data more effectively.


\bibliography{acl_latex}

\newpage
\appendix

\section{Appendix}
\label{sec:appendix}

Table~\ref{tab:appendix-web-search-dataset-categories} lists the categories used in the Web Search dataset.

\paragraph{Dataset Availabilty.} We provide both the Web Search and Common Crawl datasets, and the tokenizers in an anonymous Github repository: \url{https://github.com/frank7li/Document_Classification_using_File_Names}.

\begin{table}[]
    \centering\scalebox{0.8}{
    \begin{tabular}{l}
    \toprule
        \textbf{Category}\\
    \midrule
      \texttt{specification}   \\
      \texttt{resume}\\
      \texttt{restaurant\_menu} \\
      \texttt{price\_list} \\
      \texttt{press\_release}  \\
      \texttt{policy}  \\
      \texttt{newsletter}  \\
      \texttt{meeting\_minutes}  \\
      \texttt{map}  \\
      \texttt{letter}  \\
      \texttt{job\_posting} \\
      \texttt{guide}  \\
      \texttt{form} \\
      \texttt{course\_syllabus} \\
      \texttt{bill\_act\_law} \\
    \bottomrule
    \end{tabular}}
    \caption{Document categories used in the Web Search dataset.}
    \label{tab:appendix-web-search-dataset-categories}
\end{table}


\end{document}